\def\year{2021}\relax
\documentclass[letterpaper]{article} 
\usepackage{aaai22}  
\usepackage{times}  
\usepackage{helvet}  
\usepackage{courier}  
\usepackage[hyphens]{url}  
\usepackage{graphicx} 
\usepackage{natbib}  
\usepackage{caption} 
\DeclareCaptionStyle{ruled}{labelfont=normalfont,labelsep=colon,strut=off} 
\usepackage{algorithm}
\usepackage{algorithmic}
\usepackage{bm}
\usepackage{booktabs} 
\usepackage{amsmath}
\usepackage{makecell}
\usepackage{subfigure}
\usepackage{amssymb}
\usepackage{enumitem}
\usepackage{multicol}
\usepackage{multirow}
\usepackage{footnote}
\usepackage{natbib}
\usepackage{tikz}
\makesavenoteenv{tabular}
\usepackage{latexsym}
\usepackage[utf8]{inputenc}
\usepackage{microtype}
\usepackage{color,xcolor}
\newcommand{\tabincell}[2]{\begin{tabular}{@{}#1@{}}#2\end{tabular}}  
\definecolor{mypink}{RGB}{205,85,85}
\definecolor{myblue}{RGB}{58,95,205}
\definecolor{myorange}{RGB}{200,109,49}
\definecolor{mygrey}{RGB}{140,140,140}

\newcommand{\nop}[1]{}
\newcommand{\lb}{\mbox{$\langle$}}
\newcommand{\rb}{\mbox{$\rangle$}}

\title{Fine-Grained Opinion Summarization with Minimal Supervision}
\usepackage{bibentry}

\author{Suyu Ge, Jiaxin Huang, Yu Meng, Sharon Wang, Jiawei Han\\}

\affiliations {
  University of Illinois Urbana-Champaign \\
  {\{suyuge2, jiaxinh3, yumeng5, sjwang3, hanj\}@illinois.edu}\\}

\begin{document}
\maketitle
\begin{abstract}
\nop{
Opinion Summarization aims to profile a target by extracting opinions from multiple documents into particular aspects and sentiments.
Most existing work approaches the task in a semi-supervised manner due to the difficulty of obtaining high-quality annotation from thousands of documents.
In this work, we propose to study this problem with minimal supervision, where only aspect and sentiment names and a few keywords are available.
To face the heterogeneous nature of user opinions, we propose to first automatically identify opinion phrases from the raw corpus and classify them into different aspects and sentiments, then construct multiple fine-grained opinion clusters under each aspect/sentiment.
Each cluster consists of semantically coherent phrases, expressing uniform opinions towards certain sub-aspect or characteristics, e.g., positive feelings for ``hamburgers'' in the ``food'' aspect.
We accomplish this by (1) training an opinion-oriented spherical word embedding space to provide weak supervision for the phrase classifier, and (2) performing phrase clustering using the aspect-aware contextualized embedding generated from the phrase classifier.
Both automatic evaluations on the benchmark and quantitative human evaluation validate the effectiveness of our approach.
}
Opinion summarization aims to profile a target by extracting opinions from multiple documents.
Most existing work approaches the task in a semi-supervised manner due to the difficulty of obtaining high-quality annotation from thousands of documents.
Among them, some uses aspect and sentiment analysis as a proxy for identifying opinions.
In this work, we propose a new framework, \emph{FineSum}, which advances this frontier in three aspects: 
(1) \emph{minimal supervision}, where only aspect names and a few aspect/sentiment keywords are available; 
(2) \emph{fine-grained opinion analysis}, where sentiment analysis drills down to the sub-aspect level; and
(3) \emph{phrase-based summarization}, where opinion  is summarized in the form of phrases.
FineSum automatically identifies opinion phrases from the raw corpus, classifies them into different aspects and sentiments, and constructs multiple fine-grained opinion clusters under each aspect/sentiment.
Each cluster consists of semantically coherent phrases, expressing uniform opinions towards certain sub-aspect or characteristics (e.g., positive feelings for ``burgers'' in the ``food'' aspect).
An opinion-oriented spherical word embedding space is trained to provide weak supervision for the phrase classifier, and phrase clustering is performed using the aspect-aware contextualized embedding generated from the phrase classifier.
Both automatic evaluation on the benchmark and quantitative human evaluation validate the effectiveness of our approach.
\end{abstract}

\section{Introduction}

Opinion summarization is the task of aggregating user opinions towards a single target from multiple documents (e.g., profiling a restaurant from online reviews).
It benefits intelligent decision making by succinctly displaying diverse opinions to users and reducing the information overload.

Different from generic multi-document summarization, the large volumes of reviews and the inherent subjectivity 
within them pose challenges to curating golden annotation for this task, rendering end-to-end training infeasible.
A majority of work focuses on developing weakly-supervised or unsupervised summarization approaches~\cite{elsahar2020self,suhara2020opiniondigest}.
To further handle the diversity and conflicts in user opinions, some approaches perform aspect extraction and sentiment polarization at first, then generate summaries for different aspects and sentiments in either extractive or abstractive forms~\cite{angelidis2018summarizing,krishna2018generating}.

\begin{figure}[ht]
	\centering
	\resizebox{0.47\textwidth}{!}{\includegraphics{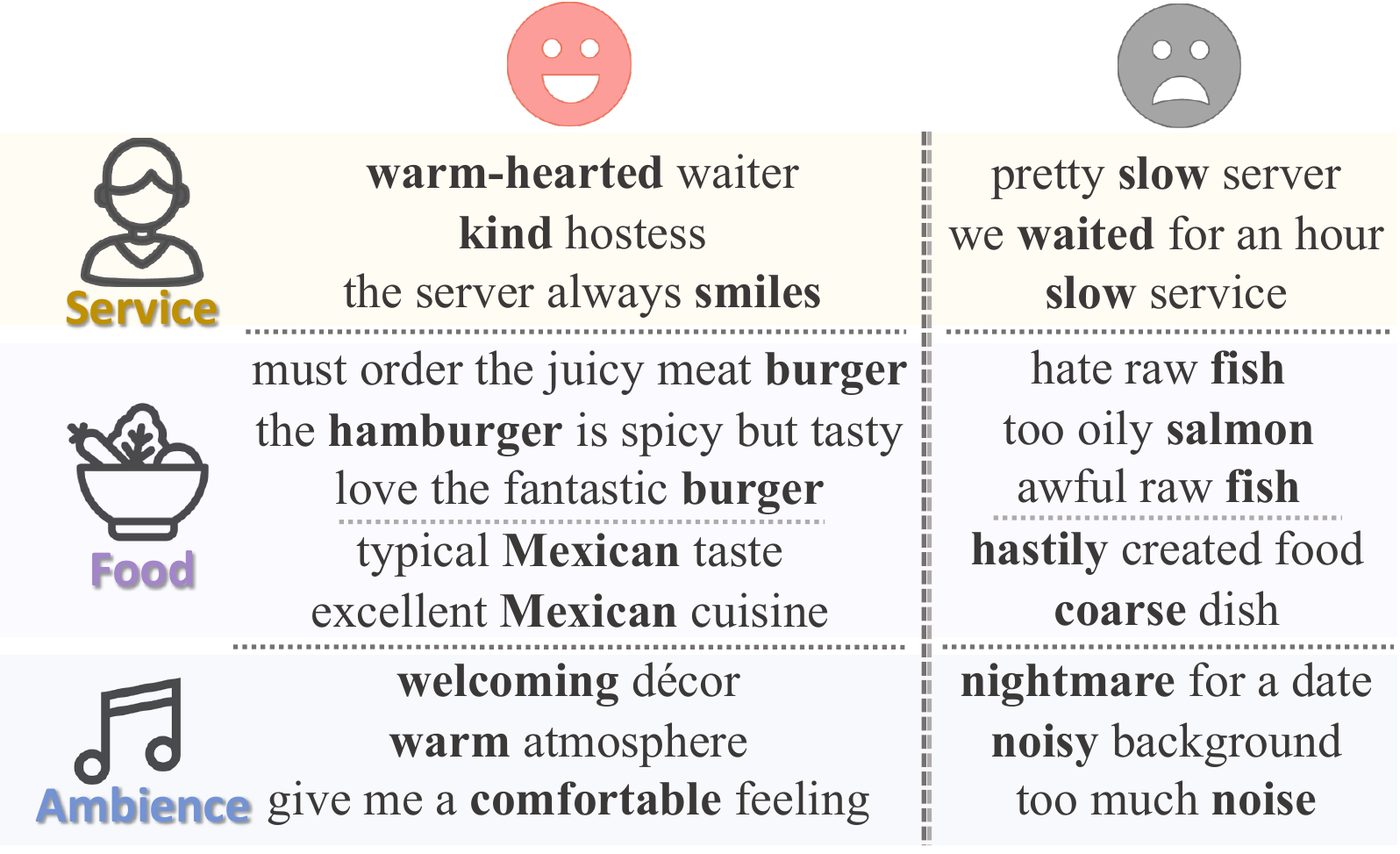}}
	\caption{An example of fine-grained opinion summarization (Opinion clusters are separated by dot lines).}
	\label{fig:task_display}
\end{figure}

Though previous methods partly consider the heterogeneity in user opinions, we argue that they still summarize at a coarse level for two reasons:
(1) Opinions in the same \lb aspect, sentiment\rb\ category may target at different subjects (e.g., in the \lb food, good\rb\ category of Fig.~\ref{fig:task_display}, one set of opinions can be about ``burger'' but the other about ``fish''), and
(2) different opinions may focus on different characteristics of the same subject (e.g., one may praise waiters for their kind service but the other may comment on their slowness).
Motivated by this, we propose to drill down and aggregate similar opinions for each sub-aspect and characteristic.

\nop{
Further, traditional extractive and abstractive methods may be sub-optimal for fine-grained summarization.
To explain, it is common to see multiple fine-grained opinions entangled in the same sentence.
In this scenario, extractive summarization will either bring outlier or cause information loss.
On the other hand, abstractive summarization usually suffers from hallucinations and text degeneration.
Therefore, instead of extracting or generating sentences, we seek to extract and aggregate smaller semantic units, such as phrases, as our fine-grained summarization.
}

Further, traditional extractive and abstractive methods may be sub-optimal for fine-grained summarization.
It is common to see multiple fine-grained opinions entangled in the same sentence.
In this scenario, extractive summarization will either bring outlier or cause information loss; whereas abstractive summarization usually suffers from hallucinations~\cite{maynez2020faithfulness} and text degeneration~\cite{holtzman2019curious}.
Therefore, instead of extracting or generating sentences, we seek to extract and aggregate smaller semantic units, such as phrases, as our fine-grained summarization.

In this paper, we propose FineSum, a fine-grained opinion summarization approach, which leverages only aspect names and a few related keywords as supervision.
It consists of the following three stages:
(1) \emph{Extract meaningful candidate phrases} by performing syntactic analysis towards the raw corpus.
(2) \emph{Classify each candidate phrase} into a particular aspect and sentiment.
We leverage two modules as phrase classifiers: (i) the opinion-oriented spherical embedding and (ii) the contextualized BERT classifier.
The opinion-oriented embedding explicitly represents aspects/sentiments along with word semantics in a distinctive sphere space, and classifies phrases according to their directional similarities with the aspect/sentiment embedding.
Meanwhile, it outputs aspect-related and sentiment-related sentences as weak supervision to train the BERT classifier.
To enhance model performance, we additionally fine-tune BERT with jointly agreed phrases from the two classifiers.
(3) \emph{Aggregate phrases within each aspect and sentiment to obtain fine-grained opinion clusters} so that phrases in the same cluster convey coherent meanings.

We summarize our contributions as follows:
(1) a minimally supervised approach is proposed for fine-grained opinion summarization, which relies only on aspect names and a few keywords;
(2) extracted candidate phrases are classified by training an opinion-oriented spherical embedding and leveraging it to provide weak supervision for a contextualized classifier, and
similar phrases are then aggregated within each aspect and sentiment to obtain fine-grained opinion clusters; and
(3) Extensive automatic and human evaluations verify the superiority of this approach.
\section{Related Work}
As the primary goal is to reduce redundancy, previous work on unsupervised or weakly-supervised opinion summarization mainly adopts a popularity-based approach (i.e., extract or generate sentences containing the most salient opinions in the original corpus)~\cite{di2014hybrid,ganesan2010opinosis}.
A majority of early methods focus on extractive summarization.
\citet{ku2006opinion} define popular opinions using TF-IDF and use pre-defined keyword sets to retrieve the most relevant and opinionated sentences.
\citet{paul2010summarizing} extract opinions according to a variety of lexical and syntactic features, and calculate salience and contrastiveness of sentences using random walk.
Recently, with the proliferation of end-to-end training, abstractive summarization receives much more attention.
A typical practice is to encode salient information using an aggregated representation, then generate novel sentences by reconstructing this representation.
A representative method is MeanSum~\cite{chu2019meansum}, which generates summaries by training an auto-decoder to reconstruct the averaged input representation.
Similarly, \citet{amplayo2021informative} condense review documents into multiple dense vectors and use a multi-source fusion module to generate summaries.
Though effective, extractive and abstractive summarizations suffer from information loss since they only consider popular opinions.
Moreover, they overlook the heterogeneity and conflicts in opinions by generating uniform summaries for all types of opinions.

Only a few unsupervised methods propose to summarize according to different aspects and sentiments~\cite{krishna2018generating,frermann-klementiev-2019-inducing}.
\citet{angelidis2018summarizing} first couple extractive opinion summarization with the tasks of aspect identification and sentiment polarization.
It uses an aspect extractor trained under a multi-task objective and a sentiment predictor based on multiple instance learning.
Following this, \citet{angelidis2021extractive} represent aspects as discrete latent codes in the quantized transformer space, and encode sentences to the aspect space using a variational autoencoder.
However, the above methods remain coarse-grained because they are not designed to capture sub-aspects and specific characteristics within each aspect.
On the contrary, we propose to generate fine-grained phrase clusters for sub-aspects.
To our knowledge, this is the first work attempting to generate opinion summarization in the form of phrase clusters.
\begin{figure*}[t]
	\centering
	\resizebox{0.99\textwidth}{!}{\includegraphics{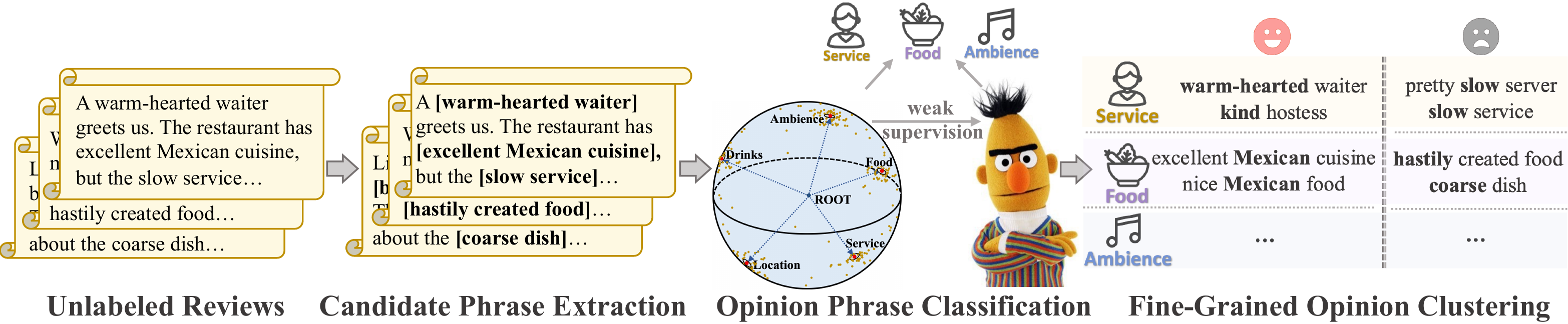}}
	\caption{An overview of FineSum: Our proposed fine-grained opinion summarization approach.}
	\label{fig:approach}
\end{figure*}

\section{Problem Definition}
Given a corpus $T$ containing reviews for targets $\{t_1, t_2, \ldots\}$ from a single domain (e.g., restaurants), we define a domain-related aspect set $A=\{a_1, a_2, \ldots\}$ and sentiment set $S=\{s_1, s_2, \ldots\}$ and input a keyword list as $L_a$ or $L_s$ for each $a$ or $s$. 
For every target $t$, we define its review sentence set as $R = \{r_1,r_2, \ldots\}$, where each review $r$ consists of multiple sentences $(x_1, x_2, \ldots)$.
Each phrase $p$ is defined as a non-overlapped word sequence $(w_1, w_2, \ldots)$ in one sentence.
For each target, our final model outputs are a set of clusters $C=\{c_1, c_2, \ldots\}$ for every aspect-sentiment pair \lb $a, s$\rb, where each cluster $c$ contains multiple semantically coherent phrases $(p_1, p_2, \ldots)$.

\section{Approach}
Fig.~\ref{fig:approach} illustrates the overall workflow of our approach, which decomposes the task of aspect-based fine-grained opinion summarization to three stages.
The first stage, \textit{candidate phrase extraction}, performs syntactic analysis to bring up consecutive and informative multi-word sequences from raw corpus as candidate phrases (Section~\ref{sec:extraction}). 
The following stage, \textit{opinion phrase classification}, aims to classify extracted phrases into different aspects and sentiments (Section~\ref{sec:classification}).
The last  stage, \textit{opinion phrase clustering}, generates fine-grained clusters within each aspect and sentiment by gathering semantically coherent phrases into the same cluster (Section~\ref{sec:clustering}).
We introduce them as follows.

\subsection{Candidate Phrase Extraction}
\label{sec:extraction}
\begin{figure}[ht]
	\centering
	\resizebox{0.47\textwidth}{!}{\includegraphics{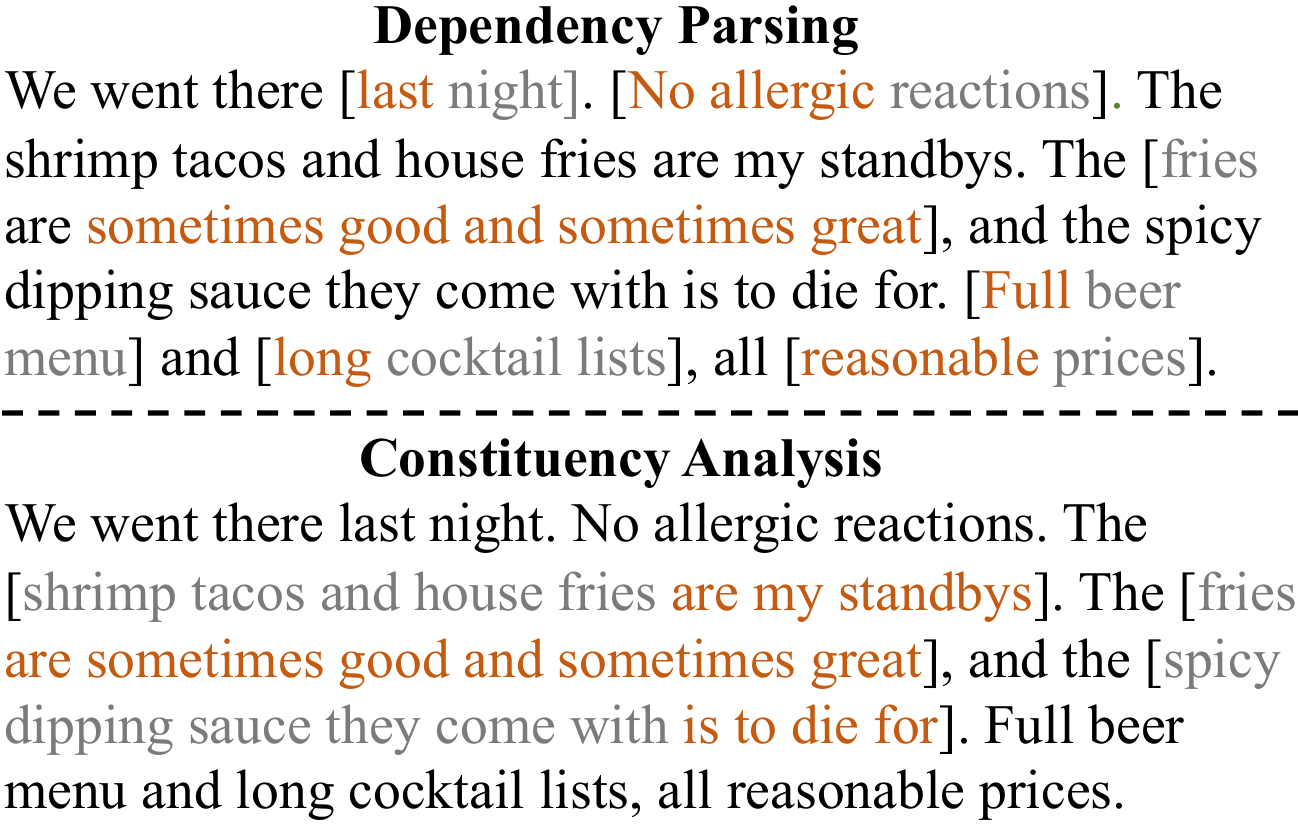}}
	\caption{An illustration of phrases extracted by dependency parsing and constituency analysis.
	We highlight the \textcolor{mygrey}{subject} and its \textcolor{myorange}{associated descriptions} inside each [phrase].}
	\label{fig:phrase_mine}
\end{figure}
As phrases form concise but complete semantic units in the original review sentences, we propose to extract phrases as the basic component of summarization, instead of the whole sentences.
However, existing phrase mining methods are usually designed to extract entity-like structures, which differs from our goal of extracting opinions. 

To mine opinions, we need to simultaneously discover a \textbf{\textit{subject}}, which is usually in the \textbf{\textit{noun}} form, along with its associated \textbf{\textit{descriptions}}, which are usually in the form of \textbf{\textit{adjective}}, \textbf{\textit{adverb}}, or \textbf{\textit{verb}}.
We achieve this by performing syntactic analysis towards each sentence using custom parsing tools.
Specifically, we consider two types of syntactic structure:
(1) \textbf{dependency parsing} which discovers a \textit{noun} subject and its descriptive \textit{adjectives}/\textit{adverbs}:
If a \textit{noun} is in one-hop relation with any \textit{adjective} or \textit{adverb}, we count them together as a phrase; and
(2) \textbf{constituency analysis} which identifies a \textit{noun} subject and its consecutive \textit{verb} phrases:
If a \textit{noun} component and a \textit{verb} component exist at the same level in the constituency parsing tree, they are considered as a phrase.
We illustrate the results of both methods in Fig.~\ref{fig:phrase_mine}.
More technical details of it can be found in the Appendix.
From the figure, dependency parsing usually mines concise phrases with similar structures, whereas constituency analysis can identify phrases in more diverse expression forms.
We take the union of their results as our final candidate phrase set, resulting in a phrase extractor with high recall.
According to our human evaluation on 50 reviews, the recall of this method reaches 0.92, suggesting it as an intuitive yet effective approach.
Note that this semantic-agnostic method will inevitably introduce noisy phrases such as ``last night''.
However, this will be corrected by the subsequent classification stage.
At this stage, our goal is to bring up as many candidates as possible.

\subsection{Opinion Phrase Classification}
\label{sec:classification}
\paragraph{Learning opinion-oriented embedding.}
Given the aspect/sentiment set and keyword lists, we leverage them as seeds to learn text embeddings tailored for aspects/sentiments.
For simplicity, 
we take aspect categories as an example to introduce our phrase embedding learning and classification method, and sentiment categories follow the same procedure.
To learn an aspect-distinctive embedding space,
we jointly embed text and aspects in the spherical space, where directional similarity is used to effectively characterize semantic correlations among words, sentences, and aspects.
Each aspect is surrounded by its representative keywords on the sphere.
In this way, the aspect information of a word is explicitly measured by its directional similarity with different aspects.

The opinion-oriented spherical embedding is implemented via hierarchical topic mining~\cite{meng2020hierarchical}. We simplify it to the flat case and the learning objectives are introduced as below:

\vspace{-1.5ex}
{\small
\begin{equation} 
\label{eqn2}
\begin{aligned}
    \forall a, x, &w,\quad ||\mathbf{a}||=||\mathbf{x}||=||\mathbf{w}||=1,\\
    \mathcal{L}_{inter}=\sum_{a_i\in A}& \sum_{a_j\in A \setminus \{a_i\}} min(0,1-\textbf{a}_i^T\textbf{a}_j-m_{inter}).\\
    \mathcal{L}_{intra}=&\sum_{a_i\in A} \sum_{w_j\in L_{a_i}} min(0,\mathbf{w}_j^T\textbf{a}_i-m_{intra}).\\
    \mathcal{L}_{aspect}=&\sum_{x\in T} \log p(\mathbf{x}|\mathbf{a}_x)+\\
    &\sum_{x\in T} \sum_{w_i\in x} \log p(\mathbf{w}_i|\mathbf{x})+\\
    &\sum_{x\in T} \sum_{w_i\in x} \sum_{|j|\leq h,j\notin 0} \log p(\mathbf{w}_{i+j}|\mathbf{w}_{i}).\\
    \mathcal{L}=&-\mathcal{L}_{inter}-\mathcal{L}_{intra}-\mathcal{L}_{aspect},
\end{aligned}
\end{equation}
}
\noindent
where $m_{inter}$ and $m_{intra}$ are two learnable parameters, and $h$ is
the context window length.
The first objective $\mathcal{L}_{inter}$ encourages inter-aspect distinctiveness across different aspects by enforcing the cosine distance between any two aspects to be larger than $m_{inter}$.
The second objective $\mathcal{L}_{intra}$ requires the embeddings of aspect keywords to be placed near the aspect center direction within a local region $m_{intra}$.
The third objective $\mathcal{L}_{aspect}$ models the corpus generation process conditioned on the aspects in a three-step process:
(1) $p(\mathbf{x}|\mathbf{a}_x)$ conditions each sentence $\mathbf{x}$ on an aspect $\mathbf{a}_x$,
(2) $p(\mathbf{w}_i|\mathbf{x})$ models the semantic coherence between a word $\mathbf{w}_i$ and the sentence $\mathbf{x}$ it appears, and
(3) $p(\mathbf{w}_{i+j}|\mathbf{w}_{i})$ models co-occurring words within local contexts.
Note that all three steps use directional similarity to model correlations.
Namely, $p(\mathbf{x}|\mathbf{a}_x) \propto \mathbf{x}^T\mathbf{a}_x$.
More details on modeling and optimization can be found in the original topic mining paper.

\paragraph{Knowledge distillation to contextualized classifier.}
Context is crucial for phrase aspect classification. 
The context-free spherical embedding space mainly captures word-level discriminative signals but is insufficient to model sequential information from ordering of words.
Therefore, we propose to distill knowledge from the aspect-regularized embedding space by generating confident soft predictions for high-quality sentences, and fine-tune a pre-trained language model. 
Specifically, we leverage soft predictions given by the directional similarity between sentence embeddings and aspect embeddings as weak supervision to fine-tune the BERT-base model~\cite{devlin2019bert} for aspect classification\footnote{Similar with aspect classification, we fine-tune another BERT-base model to classify sentiments.}.

To provide high-quality supervision, for each aspect $a_i$, we select top-$K$ sentences from the entire corpus with the highest $\mathbf{x}^T\mathbf{a}_i$ score.
We transform the directional similarity to pseudo training labels $\mathbf{l}_x$ as below:

\begin{equation}
    l_{xi}=\frac{\exp ( \alpha \cdot \mathbf{x}^T\mathbf{a}_i)}{\sum\limits_{a_i \in A} \exp ( \alpha \cdot \mathbf{x}^T\mathbf{a}_i)},
\end{equation}
where $l_{xi}$ is the probability of sentence $x$ belonging to the $i_{th}$ aspect. $\alpha$ is the temperature to control how greedy we want to learn from the embedding-based prediction.
We then train BERT on the pseudo training sentences by minimizing the cross entropy $H$ between the embedding-based prediction $\mathbf{l}$ and the output prediction $\mathbf{y}$ of BERT, namely

\begin{equation}
    H(x)=\sum_x \sum_i l_{xi} \log \frac{l_{xi}}{y_{xi}}.
\end{equation}

\paragraph{Fine-tuning classifier on jointly agreed phrases.}
To identify the aspect of candidate phrases, one direct method is to leverage the sentence-level classifier fine-tuned in previous stage.
However, the sentence-level model may not perform as well on phrases because phrases are usually shorter, thus provide less context for accurate classification.
Thus, we propose to further enhance the BERT model by fine-tuning it for phrase-level aspect classification.
Note that the candidate phrase extraction stage may include phrases not belonging to any aspect.
To exclude them, we additionally require the model to identify such ``background'' phrases by training BERT to output a uniform aspect distribution on them.

To generate high-quality pseudo training samples for BERT, we exploit the wisdom from both models (i.e., the opinion-oriented embedding and the weakly trained BERT). 
We select their \textit{jointly agreed predictions} as pseudo training phrases, and generate pseudo labels $l'_{s}$ for phrase $s$ as:

\begin{equation}
l'_{si}=\left\{
\begin{array}{ll}
\frac{\exp ( \alpha y_{si}) }{\sum\limits_{i \in |A|} \exp ( \alpha y_{si})} & {y_{si}\geq \theta_1, \overline{\mathbf{w}}^T_s\mathbf{a}_i \geq \theta_2}\\
\frac{1}{|A|} & {y_{si}< \theta_1, \overline{\mathbf{w}}^T_s\mathbf{a}_i< \theta_2}
\end{array}, \right.
\end{equation}
where $\overline{\mathbf{w}}_s$ is the averaged embeddings of words in the phrase $s$, $y_{si}$ is the predicted probability from BERT, and $\overline{\mathbf{w}}^T_s\mathbf{a}_i$ is the directional similarity between phrase $s$ and the $i^{th}$ aspect in the embedding space.
$\theta_1$ and $\theta_2$ are two probability thresholds.
During inference, we also use $\theta_2$ for BERT to decide whether a phrase belongs to a certain aspect.

\subsection{Fine-Grained Opinion Clustering}
\label{sec:clustering}
Given predictions from the previous stage, we can organize all extracted phrases according to their aspects and sentiments.
However, there are two problems with this practice:
(1) Phrases located in the same aspect and sentiment may still cover diverse and heterogeneous opinions, varied by their subjects and targeted characteristics; and
(2) some phrases express similar meanings using different words, which leads to unwanted semantic redundancy.
To solve these problems, we propose to represent fine-grained opinions by automatically forming clusters under each aspect and sentiment.
To guarantee that phrases belonging to the same cluster convey consistent and coherent meanings, we require them to locate near each other in the semantic space.
We achieve this by clustering in the fine-tuned BERT embedding space, as it explicitly encodes aspect and sentiment information after training and fine-tuning in the previous stage.

Specifically, we use the bottom-up hierarchical agglomerative clustering~\cite{dubitzky2013encyclopedia}
which treats each phrase as a singleton cluster at the outset, and then successively merges pairs of phrases until the euclidean distance between phrases in the same cluster exceeds a pre-defined threshold $T_c$.


The final output of \textit{FineSum} is the fine-grained clusters under each aspect and sentiment.
Considering the semantic coherence in each cluster, one could further reduce redundancy by selecting salient phrase(s) to represent the whole cluster, or even selecting salient cluster(s) to represent each aspect-sentiment pair.
In this work, we stay on the original complete clusters to offer comprehensive summarization.
In real application, the best form of summarization should be adjusted according to user needs, which is beyond the scope of this paper.

\section{Experiments}
As no off-the-shelf evaluation framework exists, we evaluate model performance quantitatively and qualitatively on two major tasks in our approach: (i) opinion phrase classification and (ii) fine-grained opinion clustering.
For each task, we create extra human annotation and use them for performance analysis.
We will release the source code and test data together with human-curated annotations.

\begin{table}[t]
	\centering
	\resizebox{0.48\textwidth}{!}{
	\begin{tabular}{|c|c|c|c|}
    \hline
	\multirow{4}{*}{Restaurant}&\textbf{\# Training Sentences} & \textbf{\# Training Phrases} & \textbf{\# Test Reviews}\\
	\cline{2-4}
    &10,000 &297,210 &643\\
    \cline{2-4}
    &\multicolumn{2}{|c|}{\textbf{Aspects}} & \textbf{Sentiments}\\
    \cline{2-4}
    &\multicolumn{2}{|c|}{location, drinks, food, ambience, service}&good, bad\\
    \hline
    \multirow{4}{*}{Laptop}&\textbf{\# Training Sentences} & \textbf{\# Training Phrases} & \textbf{\# Test Reviews}\\
	\cline{2-4}
    &16,000 &83,540 &307\\
    \cline{2-4}
    &\multicolumn{2}{|c|}{\textbf{Aspects}} & \textbf{Sentiments}\\
    \cline{2-4}
    &\multicolumn{2}{|c|}{\tabincell{c}{support, os, display, battery, company,\\ mouse, software, keyboard}}&good, bad\\
    \hline
	\end{tabular}}
	\caption{Dataset Statistics.}
	\label{tab:dataset}
\end{table}

\begin{table*}[t]
	\centering
	\resizebox{\textwidth}{!}{
	\begin{tabular}{c|cccccccccccccccc}
    \Xhline{1.3pt}
	\multirow{2}{*}{\textbf{Model}}&\multicolumn{4}{c|}{\textbf{Restaurant-aspect}}&\multicolumn{4}{c|}{\textbf{Restaurant-sentiment}}&\multicolumn{4}{c|}{\textbf{Laptop-aspect}}&\multicolumn{4}{c}{\textbf{Laptop-sentiment}}\\
            &\multicolumn{1}{c}{Acc.} & \multicolumn{1}{c}{Pre.} & \multicolumn{1}{c}{Rec.} & \multicolumn{1}{c|}{{\scriptsize macro-}F1} &
            \multicolumn{1}{c}{Acc.} & \multicolumn{1}{c}{Pre.} & \multicolumn{1}{c}{Rec.} & \multicolumn{1}{c|}{{\scriptsize macro-}F1} &
            \multicolumn{1}{c}{Acc.} & \multicolumn{1}{c}{Pre.} & \multicolumn{1}{c}{Rec.} & \multicolumn{1}{c|}{{\scriptsize macro-}F1}&
            \multicolumn{1}{c}{Acc.} & \multicolumn{1}{c}{Pre.} & \multicolumn{1}{c}{Rec.} & \multicolumn{1}{c}{{\scriptsize macro-}F1} \\
			\hline
			CosSim & 64.20& 54.55& 47.82& 49.85& 70.19& 68.73& 63.57& 63.87&55.39&59.21&58.72&54.33&70.03&69.88&70.50&70.84\\
			W2VLDA* &71.05 &61.98 &61.46 &54.30 &77.36 &79.69 &73.02 &70.30&66.21&67.83&68.45&65.02&71.69&71.33&71.51&71.66\\
			BERT &75.98 &62.30 &\textbf{80.26} &60.83 &79.50 &77.72 &77.99 &77.85&69.12&68.13&68.74&67.38&71.24&70.92&71.08&71.17\\
			JASen* &84.99 &\textbf{74.09} &77.50 &\textbf{74.85} &83.64 &83.45 &80.76 &81.74&73.02&70.31&70.67&72.19&76.47&76.67&76.46&76.53\\
			\hline
			FineSum w/o BERT &81.18 &71.31 &61.53 &64.52 &67.19 &75.66 &56.55 &51.96&67.34&68.42&68.71&67.43&68.12&67.94&69.47&68.33\\
			FineSum w/o joint &84.91 &69.72 &76.74 &68.89 &84.76 &84.92 &83.94 &84.10&71.01&72.18&71.27&70.42&78.83&78.38&78.74&78.82\\
			FineSum &\textbf{87.67} &69.48 &78.31 &69.90 &\textbf{86.16} &\textbf{87.21} &\textbf{85.82} &\textbf{84.86}&\textbf{76.98}&\textbf{76.27}&\textbf{76.96}&\textbf{75.93}&\textbf{79.15}&\textbf{79.64}&\textbf{79.02}&\textbf{79.15}\\
    \Xhline{1.3pt}
	\end{tabular}}
	\caption{Quantitative evaluation of aspect identification and sentiment polarity on sentence level tasks.
	* denotes that the model learns aspect and sentiment jointly. }
	\label{tab:sent_classify}
\end{table*}

\begin{table}[t]
	\centering
	\resizebox{0.43\textwidth}{!}{
	\begin{tabular}{c|cccc}
    \Xhline{1pt}
	\multirow{2}{*}{\textbf{Model}}&\multicolumn{4}{c}{\textbf{Restaurant-phrase}}\\
            &\multicolumn{1}{c}{Acc.} & \multicolumn{1}{c}{Pre.} & \multicolumn{1}{c}{Rec.} & \multicolumn{1}{c}{{\scriptsize macro-}F1}\\
			\hline
			CosSim &55.97&56.22&53.64& 53.84\\
			W2VLDA* &63.34&61.64&61.37&60.93 \\
			BERT &72.39&63.54&74.13&67.55 \\
			JASen* &82.20 &85.32 &79.70 &79.64\\
			\hline
			FineSum w/o BERT &63.80 &62.76 &60.19 &58.21\\
			FineSum w/o joint &76.82 & 66.41&78.98 &70.43\\
			FineSum &\textbf{88.60} &\textbf{85.44} &\textbf{86.98} &\textbf{84.57}\\
    \Xhline{1pt}
	\end{tabular}}
	\caption{Quantitative evaluation of restaurant aspect identification on phrase level task.}
	\label{tab:phrase_classify}
\end{table}

\subsection{Datasets and Experimental Settings}
\subsubsection{Datasets.}
We experiment on reviews from two domains: restaurant and laptop.
Details of dataset statistics can be found in Table~\ref{tab:dataset}.
We additionally provide four seed keywords for each aspect and sentiment as weak supervision, as listed in the Appendix.
We also display the robustness of our methods on seed keywords in Section~\ref{sec:seeds}.

\noindent\textbf{Restaurant:}
We collect restaurant reviews from the Yelp Dataset Challenge\footnote{https://www.kaggle.com/yelp-dataset/yelp-dataset} as our in-domain training and summarization corpus.
We gather reviews from 42 businesses, where each business has at least 100 reviews and includes the keyword ``restaurant'' in its meta business type list.
The average number of reviews for each restaurant is 599.
For evaluation, we use an external benchmark dataset in the restaurant domain of SemEval-2016~\cite{pontiki-etal-2016-semeval}, which provides sentence-level aspect category and sentiment polarity of each review.
Following Huang et al.~\shortcite{huang2020aspect}, we remove sentences with multiple labels or with a neutral sentiment polarity to simplify the problem.

\noindent\textbf{Laptop:}
We collect reviews of 50 laptop products from the Laptop domain of Amazon Review Dataset~\cite{ni2019justifying}.
Each product has at least 100 reviews.
We also use the laptop domain of SemEval-2016 for evaluation and remove sentences with multiple labels .

\subsubsection{Experimental Settings.}
For text prepossessing, we use NLTK tokenizers and the Stanford CoreNLP~\cite{manning2014stanford} parser\footnote{https://stanfordnlp.github.io/CoreNLP/parse.html}$^{,}$\footnote{https://stanfordnlp.github.io/CoreNLP/depparse.html}.
We set the hyperparameters as follows: $h=5$, $K=2000$, $\theta_1 = 0.35$, $\theta_2 = 0.30$, $T_c =7$, embedding dimension $=100$, learning rate $=1e-5$, batch size = $64$.
We train and fine-tune BERT with AdamW optimizer for only one epoch to avoid overfitting on noisy pseudo labels.
Details on hyperparameter selection can be found in the Appendix.

\subsection{Opinion Classification}
We first evaluate the phrase classifier on sentence-level aspect identification and sentiment polarization tasks using the benchmark test set.
However, a strong sentence-level classifier may not perform as well on phrase-level task, so we additionally evaluate the model on phrase-level aspect classification task.
Due to the shortage of phrase-level annotation, we manually collect data by randomly sampling 500 extracted phrases from the Restaurant domain.
We ask three human annotators to label their aspects and split them by half into validation and test sets.
As the phrase extraction is aspect-agnostic, the collected set includes noisy phrases that do not belong to any aspect.
To handle this, we allow multiple and none aspect assignment.
Final labels were obtained using a majority vote among annotators.
The three annotators provided the same labels on 85.60\% phrases, guaranteeing the high quality of our annotation.
More annotation and evaluation details can be found in the Appendix.

\leftmargini=12pt
We compare our approach with a series of weakly-supervised baselines and two variants of our own approach. 
\begin{itemize}
\parskip -0.4ex
\item \textbf{CosSim}: a Word2Vec embedding based model, which classifies according to the cosine similarity between the averaged word embedding and topic vectors. Topic vectors are calculated as the average of seed keywords.
\item \textbf{W2VLDA}~\cite{garcia2018w2vlda}: an aspect-based sentiment analysis model, which leverages aspect and sentiment keywords as seeds to perform joint topic modeling.
\item \textbf{BERT}~\cite{devlin2019bert}: a language model fine-tuning model. 
We incorporate sentences containing seed keywords as pseudo training samples to fine-tune BERT-base.
    \item \textbf{JASen}~\cite{huang2020aspect}: a state-of-the-art aspect based sentiment analysis model. It first learns aspect-sentiment joint word embedding, then generalizes word knowledge to neural models through weakly-supervised training and iterative self-training.
    \item \textbf{FineSum w/o BERT}: A context-free ablation of our model.
    It uses the opinion-oriented spherical embedding as the only classifier.
    \item \textbf{FineSum w/o joint}: A BERT-based ablation without fine-tuning on the jointly agreed phrases (but trained on sentence-level classification).
\end{itemize}

\begin{table}[t]
	\centering
	\resizebox{0.48\textwidth}{!}{
	\begin{tabular}{lll}
    \Xhline{1.2pt}
            \textbf{Spherical Embedding} & \textbf{Vanilla BERT} & \textbf{BERT in FineSum}\\
			\hline
			allergic reactions &severe allergies  &severe allergies\\
			\textcolor{red}{severe allergies}&\textcolor{blue}{broken leg} &intestinalgastro issues\\
			mild reactions& severe food poisoning&severe food poisoning\\
			\textcolor{red}{no allergic reactions}&\textcolor{blue}{suffer severe migraine} &crazy allergies\\
    \Xhline{1.2pt}
	\end{tabular}}
	\caption{Qualitative evaluation of fine-grained clustering on the restaurant dataset. 
	We compare opinion clusters from different embeddings that show similar semantics.
	\textcolor{red}{Conflicting} and \textcolor{blue}{irrelevant} phrases are denoted with \textcolor{red}{red} and \textcolor{blue}{blue}.}
	\label{tab:cluster_case}
\end{table}

\begin{figure*}[t]
    \centering
\subfigure[Spherical Embedding]{\resizebox{0.32\textwidth}{!}{\includegraphics{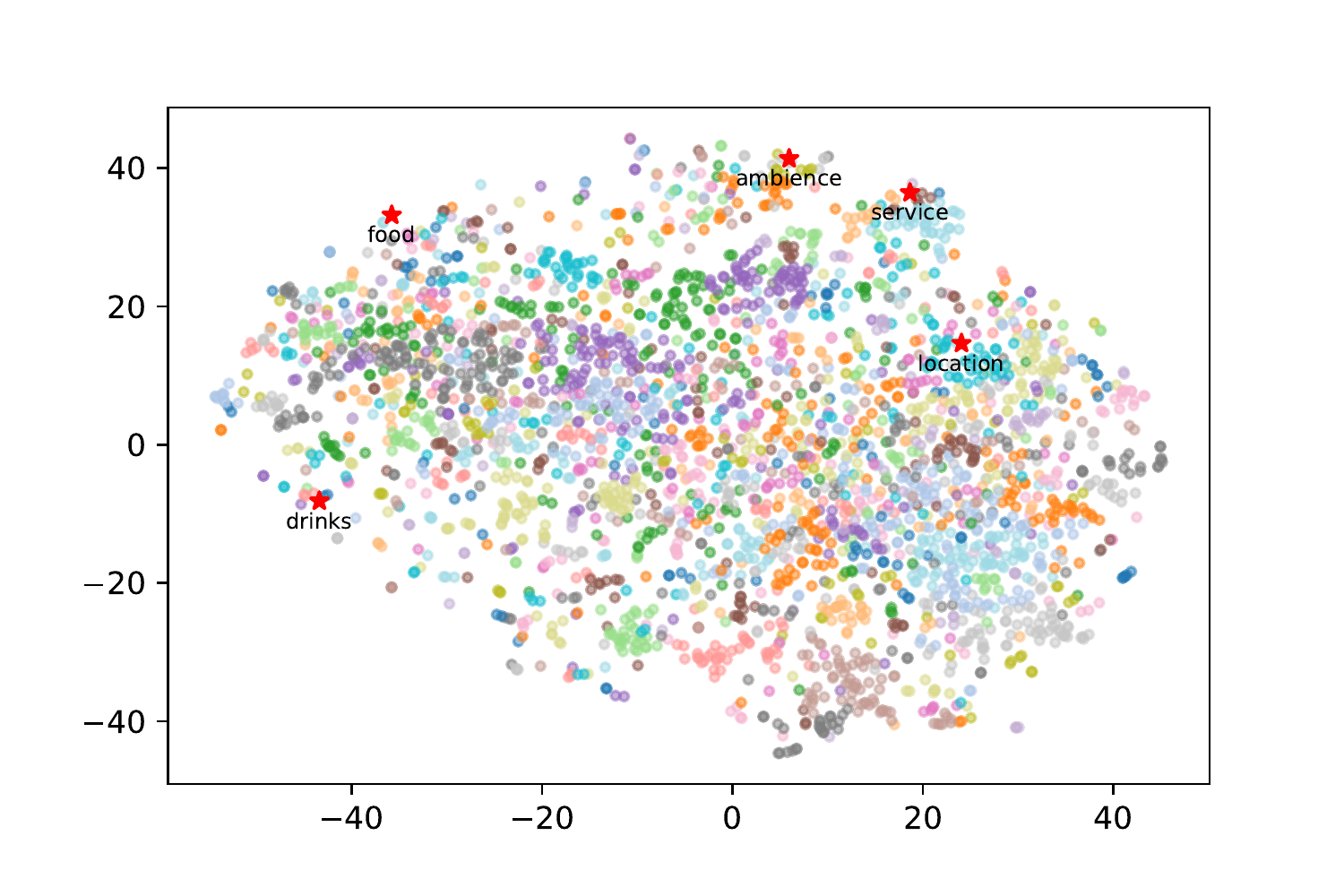}}\label{emb_josh}}
\subfigure[Vanilla BERT]{\resizebox{0.32\textwidth}{!}{\includegraphics{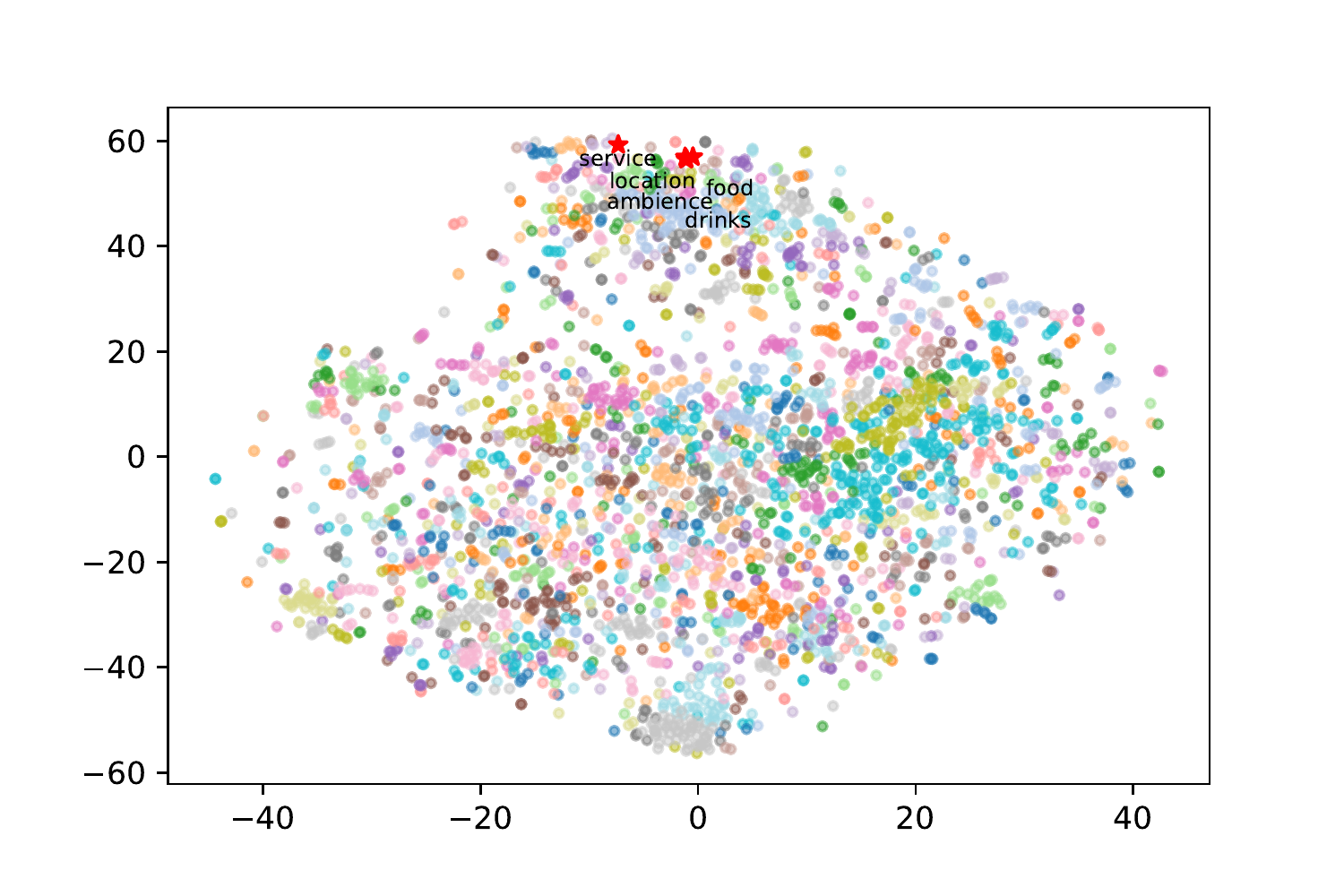}}\label{emb_bert}}
\subfigure[Fine-tuned BERT in FineSum]{\resizebox{0.32\textwidth}{!}{\includegraphics{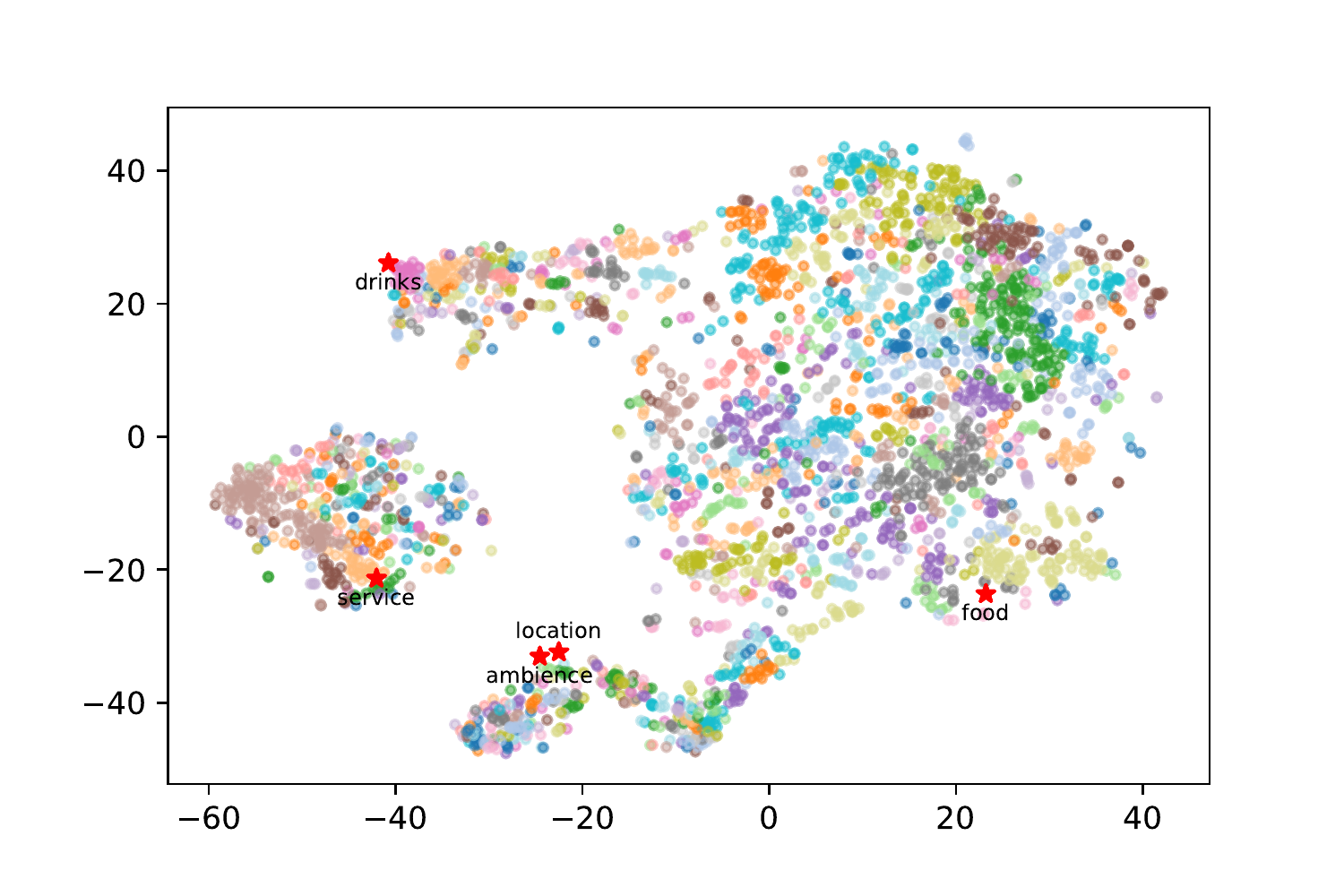}}\label{emb_bert_joint}}
\caption{Visualization of opinion clusters on Restaurant.
Phrases assigned to the same cluster are denoted with the same color.}
\label{fig:emb_vis}
\end{figure*}

The standard Accuracy, Precision, Recall and macro-F1 are used as evaluation metrics.
We run experiments for 5 times and report average performances.
For each method, we set a lowest threshold for their output classification probability.
We classify phrases with a probability larger than the threshold into the corresponding aspect, otherwise the ``none'' aspect.
The threshold is selected when the model performs best on the validation set.
Sentence- and phrase-level evaluation results are shown in Table~\ref{tab:sent_classify} and Table~\ref{tab:phrase_classify}, respectively.
We have the following observations:

\noindent\textbf{Sentence Classification.} 
(1) \textit{FineSum} outperforms all the baselines on sentiment polarization, and achieves competitive performances on aspect identification.
Benefiting from both word-level and contextualized knowledge, it achieves better performances than models only based on topic modeling (\textit{W2VLDA}), language model (\textit{BERT}), or word embedding (\textit{CosSim}, \textit{JASen}).
(2) 
\textit{FineSum} outperforms its two variants by a large margin.
This is intuitive since \textit{FineSum w/o BERT} works in a context-free manner, and \textit{FineSum w/o joint} is only trained on sentence classification task.
It may not perform as well on phrases where less contexts are presented.
(3) On restaurant aspect identification, \textit{FineSum} shows relatively lower macro-F1 and precision than \textit{JASen}.
This happens because \textit{JASen} is designed for aspect-based sentiment analysis and takes advantage of the mutual information in aspects and sentiments.
However, \textit{FineSum} views them as independent for simplicity.
In fact, we can easily extend FineSum to the joint setting by modifying the word embedding approach (e.g., substitute the spherical embedding with the embedding in JASen), which may lead to even better performance.

\noindent\textbf{Phrase Classification.} 
(1) Most baselines perform worse on phrases than on sentences.
This is because fewer informative contexts are provided in phrases, thus calling for the need of phrase-level training.
(2) Despite fewer contexts, \textit{FineSum} gains a larger edge over other baselines.
This observation validates the effectiveness of fine-tuning BERT on jointly agreed phrases.
We also qualitatively compare \textit{FineSum} with its two variants in the Appendix.

\begin{table*}[t]
	\centering
	\resizebox{0.93\textwidth}{!}{
	\begin{tabular}{ccl}
    \Xhline{1.2pt}
    \multirow{8}{*}[1ex]{\tabincell{c}{Domain:\\Restaurant\\ \\ Aspect:\\Food}}& \multirow{4}{*}[1ex]{Good} & \tabincell{l}{\textcolor{mypink}{delicious} dark \textbf{\textit{chocolate}} dipping sauce; it was a \textcolor{mypink}{good} ratio of \textbf{\textit{chocolate}} to vanilla; \textcolor{mypink}{yummy} \textbf{\textit{chocolate}}\\ brownie; just ok \textbf{\textit{chocolate}} berry dessert; \textcolor{mypink}{amazing} \textbf{\textit{chocolate}} pudding dessert }\\
    \cline{3-3}
     & & \tabincell{l}{this is a \textcolor{mypink}{fantastic} change while keeping the integrity of so many familiar \textbf{\textit{ingredients}}; all of the \textbf{\textit{ingredients}}\\ went really \textcolor{mypink}{well} together; \textcolor{mypink}{appreciate} how all the \textbf{\textit{ingredients}} come together; \textbf{\textit{everything}} was \textcolor{mypink}{well} seasoned}\\
     \cline{2-3}
     & \multirow{4}{*}[1ex]{Bad} & \tabincell{l}{left the rest of the \textbf{\textit{fish}} \textcolor{myblue}{untouched}; the \textbf{\textit{salmon}} was \textcolor{myblue}{hastily} created; the \textbf{\textit{fish}} had a \textcolor{myblue}{weird} texture; the \textbf{\textit{fish}}\\seemed a little \textcolor{myblue}{oily}; the \textbf{\textit{fish}} was \textcolor{myblue}{n't} that \textcolor{myblue}{large}; i \textcolor{myblue}{hate} raw \textbf{\textit{fish}}; \textcolor{myblue}{awful} raw \textbf{\textit{fish}}
 }\\
     \cline{3-3}
     & & \tabincell{l}{the \textbf{\textit{dish}} \textcolor{myblue}{disappointed} my companion; the \textbf{\textit{food}} was absolutely \textcolor{myblue}{underwhelming}; they were really \textcolor{myblue}{disappointed}\\ with the \textbf{\textit{dish}}; my \textcolor{myblue}{fault} for ordering their \textbf{\textit{food}}, not theirs for making it; I \textcolor{myblue}{wasn't} as impressed with the \textbf{\textit{food}}}\\
    \hline

\hline
    \multirow{8}{*}[1ex]{\tabincell{c}{Domain:\\Laptop\\ \\ Aspect:\\Mouse}} & \multirow{4}{*}[1ex]{Good} & \tabincell{l}{i \textcolor{mypink}{love} the \textbf{\textit{trackman wheel}}; the \textbf{\textit{trackman wheel}} is practically \textcolor{mypink}{perfect} just as it is; the \textbf{\textit{trackman wheel}}\\ optical is the \textcolor{mypink}{best} of those; more \textcolor{mypink}{compact} \textbf{\textit{trackman wheel}}; the \textbf{\textit{trackman wheel}} feels \textcolor{mypink}{good} in the hand
}\\
    \cline{3-3}
    & & \tabincell{l}{the two little \textbf{\textit{buttons}} are \textcolor{mypink}{useful} if you install the logitech software; love the fact that i \textcolor{mypink}{could program}\\ the \textbf{\textit{buttons}} for what i needed; the \textbf{\textit{buttons}} \textcolor{mypink}{can be programed} to perform various functions
}\\
     \cline{2-3}
     & \multirow{4}{*}[1ex]{Bad} & \tabincell{l}{sometimes the \textbf{\textit{mouse}} becomes \textcolor{myblue}{unresponsive} and \textcolor{myblue}{sluggish}; the movement of the \textbf{\textit{cursor}} with the \textbf{\textit{trackball}}\\ gets \textcolor{myblue}{slow} and \textcolor{myblue}{erratic} at times; possible nobody think the \textbf{\textit{mouse}} is too \textcolor{myblue}{slow} like me?
}\\
     \cline{3-3}
     & & \tabincell{l}{experienced \textbf{\textit{wrist/arm}} \textcolor{myblue}{pain} from mouses; my \textbf{\textit{mouse arm}} tends to be constantly \textcolor{myblue}{sore}; using a mouse 8\\ hours i began to get \textcolor{myblue}{pains} in my \textbf{\textit{wrist}}; i got \textcolor{myblue}{sick} of my optical \textbf{\textit{mouse}} \textcolor{myblue}{messing up} on me
}\\

    \Xhline{1.2pt}
	\end{tabular}}
	\caption{A case study of the final output from \textit{FineSum} on both datasets. 
	For simplicity, we display one aspect from each domain and randomly select two clusters under each aspect and sentiment.
	For a quick understanding, we manually \textit{italicize} and \textbf{bold} fine-grained sub-aspects, and highlight sentiment-indicating words with \textcolor{mypink}{red} (good) and \textcolor{myblue}{blue} (bad).
	}
	\label{tab:case_yelp}
\end{table*}

\subsection{Fine-Grained Opinion Clustering}
We compare our fine-tuned BERT embedding with two ablations on clustering performance: (i) Opinion-oriented \textit{Spherical Embedding} introduced in Section~\ref{sec:classification} and (ii) Last layer outputs from \textit{Vanilla BERT}, which are not trained on the aspect classification task.

\noindent\textbf{Qualitative Evaluation.} 
To intuitively understand the difference between embedding methods, we display semantically similar clusters from them in Table~\ref{tab:cluster_case}.
We observe that \textit{Spherical Embedding} relies excessively on overlapped surface words.
It tends to gather look-alike phrases even if their meanings are converse.
For instance, ``severe allergies'' and ``no allergic reactions'' are in the same cluster.
Besides, we also find that \textit{Vanilla BERT}, although less reliant on overlapped words, sometimes suffers from semantic drifts.
For example, ``broken leg'' and ``severe migraine'' are grouped together.
On the contrary, \textit{BERT in FineSum} forms clusters that are both coherent in meaning and diverse in expression.
We further visualize phrase distribution in Fig.~\ref{fig:emb_vis} to intuitively understand how clusters distribute in the fine-tuned BERT embedding space.
Compared with \textit{Spherical Embedding} and \textit{Vanilla BERT}, \textit{BERT in FineSum} displays a clearer and better-separated cluster space.

\noindent\textbf{Quantitative Evaluation.} 
We evaluate the \textit{coherence} and \textit{diversity} of generated clusters quantitatively.
\textit{Coherence} measures the semantic consistency of phrases within the same cluster, whereas \textit{diversity} measures how their expressions differ from each other.
In principle, a cluster of high quality should be in both high \textit{coherence} and \textit{diversity}, indicating that the model can gather phrases with similar semantic meanings regardless of their  expression forms.
\begin{figure}[h]
	\centering
	\resizebox{0.35\textwidth}{!}{\includegraphics{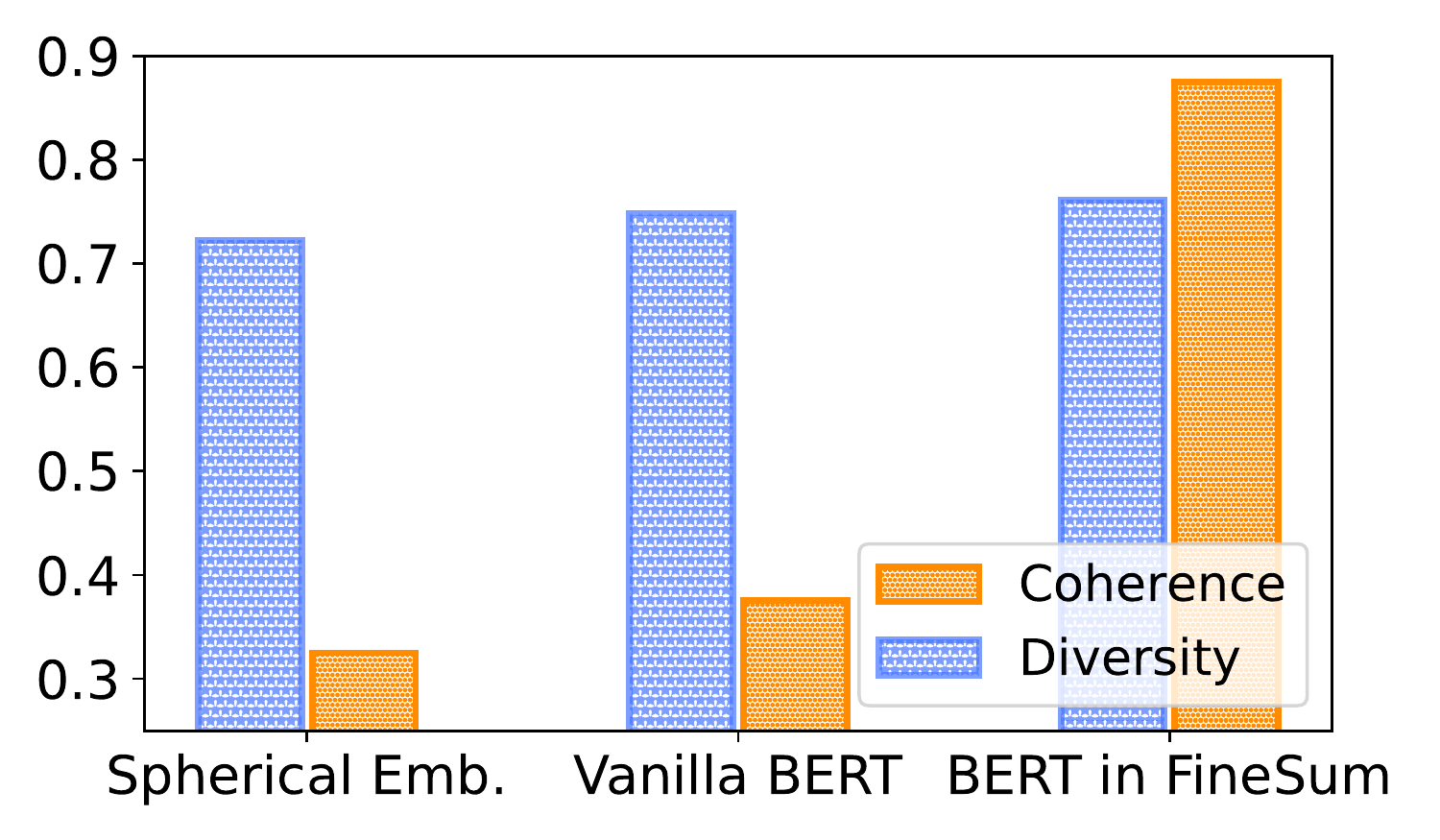}}
	\caption{Quantitative evaluation of fine-grained clustering on the restaurant dataset. Higher scores indicates better performances.}
	\label{fig:cluster_intrusion}
\end{figure}

\begin{figure}[h]
	\centering
	\resizebox{0.4\textwidth}{!}{\includegraphics{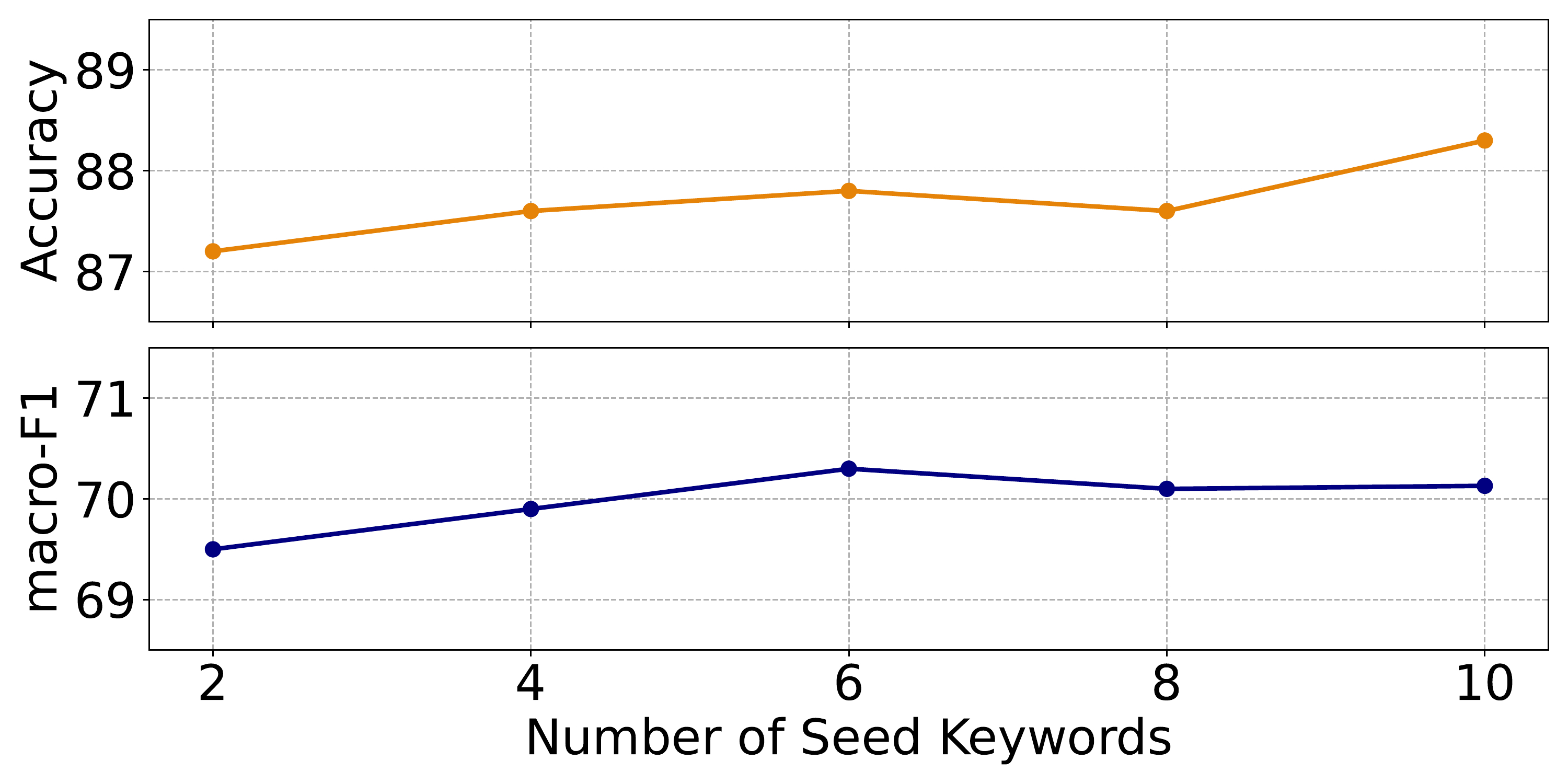}}
	\caption{Influence of seed keywords.}
	\label{fig:keyword-sensitivity}
\end{figure}
We define the two metrics as follows:
(i) \textbf{\textit{Coherence:}}  Given an opinion cluster, we inject an intrusion phrase that is randomly chosen from another cluster. 
Then three human annotators are asked to identify the intruded phrase.
Empirically, we observe that phrases within the same cluster usually share common words, making it easy to identify the intruded phrase.
Hence, we require the intruded phrase to have at least one overlapped word with other phrases.
We provide an example of this in the Appendix, along with other evaluation details.
We compute the ratio of correctly identified intrusion instances as the coherence score.
(ii) \textbf{\textit{Diversity:}} The percentage of unique words in each cluster.
Fig.~\ref{fig:cluster_intrusion} showcases that \textit{BERT in FineSum} significantly outperforms the other two ablations on \textit{coherence}, validating that the fine-tuned BERT embeddings generate semantically coherent clusters.
This finding indicates that fine-tuned BERT for aspect classification not only benefits the task itself, but also leads the model to better distinguish fine-grained sub-aspects within each aspect. 
Moreover, \textit{BERT in FineSum} achieves slightly higher \textit{diversity} score than the other two methods, indicating that the high coherence score of our approach is not brought by simply gathering phrases with similar words.

\subsection{Parameter Study}
\label{sec:seeds}
To investigate whether FineSum is sensitive to different choices of seed  keywords, we initiate the opinion-oriented embedding with different numbers of seeds. 
Figure~\ref{fig:keyword-sensitivity} shows sentence-level classification results on the Restaurant dataset.
As can be observed from the figure, the classification accuracy and macro-F1 remain relatively stable when we alter the number of seeds.
This result indicates that the opinion-oriented embedding can learn well-separated semantic space with little human guidance, which validates the robustness of FineSum and opens up possibilities to apply it to diverse domains in the future.

\subsection{Case Study}
Table~\ref{tab:case_yelp} shows an example of our system output.
We observe that different opinion phrase clusters are well-separated by their aspects and sentiments, which is guaranteed by the opinion phrase classification stage.
Probing into each aspect and sentiment, we discover that the model automatically forms clusters which represent concrete sub-aspects or describes particular traits.
For example, under the aspect-sentiment pair \lb food, bad\rb, we find one cluster expressing overall disappointment for their food and two clusters complaining about specific food types.
The coherent and meaningful clusters under each aspect validate the effectiveness of clustering with fine-tuned BERT embedding.

\section{Conclusion}
In this paper we propose \textit{FineSum}, a minimally supervised approach for fine-grained opinion summarization.
\textit{FineSum} works by first extracting candidate phrases, then classifying them into aspects and sentiments using the opinion-oriented spherical embedding and the weakly-supervised BERT.
We further propose to aggregate similar phrases using the fine-tuned BERT embedding to obtain fine-grained opinion clusters.
Comprehensive automatic and human evaluation demonstrate that our approach generates high-quality phrase-level summarization.
\clearpage

\bibliography{aaai22} 
\end{document}


\appendix
\section{Appendix}

\begin{figure}[h]
	\centering
	\resizebox{0.41\textwidth}{!}{\includegraphics{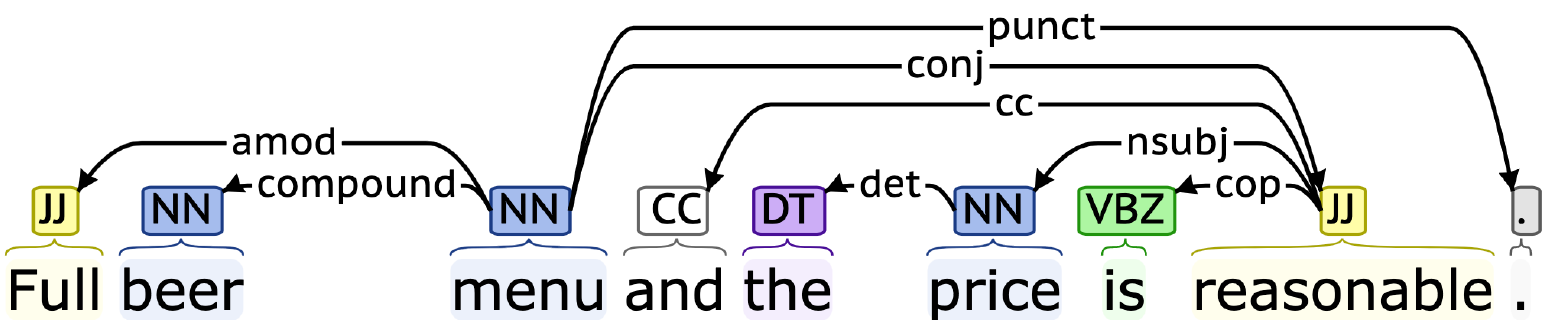}}
	\caption{An example result of dependency parsing.}
	\label{fig:dependency}
\end{figure}

\subsection{Details on Candidate Phrase Extraction}
We introduce technical details of two modules in the candidate phrase extraction stage, i.e., dependency parsing and constituency analysis.

\noindent\textbf{Dependency Parsing.} 
This module aims to discover a \textit{noun subject} and its \textit{descriptive} \textit{adjectives}/\textit{adverbs} together as a phrase.
Given a review sentence, it first recognizes all nouns, adjectives, and adverbs according to their pos-tags.
For instance, in Figure~\ref{fig:dependency}, the pos-tag ``NN'' corresponds to noun, so ``beer menu'' and ``price'' are recognized as noun (or compound noun).
Similarly, ``full'' and ``reasonable'' are adjectives.
The next step is to investigate the dependency relationship between noun subject and adjectives/adverbs.
This can be achieved by directly examining the results of our dependency parser.
As illustrated in Figure~\ref{fig:dependency}, the ``amod'' relation indicates the adjective ``full'' is used to modify the noun ``menu'', and the ``nsubj'' relation indicates the noun ``price'' is the subject of the adjective ``reasonable''.
Empirically, we use the ``amod'' and ``nsubj'' relation to decide whether a noun and an adjective is dependent.
If a noun and an adjectives/adverb are connected by a one-hop ``amod'' or ``nsubj'' relation, we count them as one candidate phrase.

\begin{figure}[t]
	\centering
	\resizebox{0.41\textwidth}{!}{\includegraphics{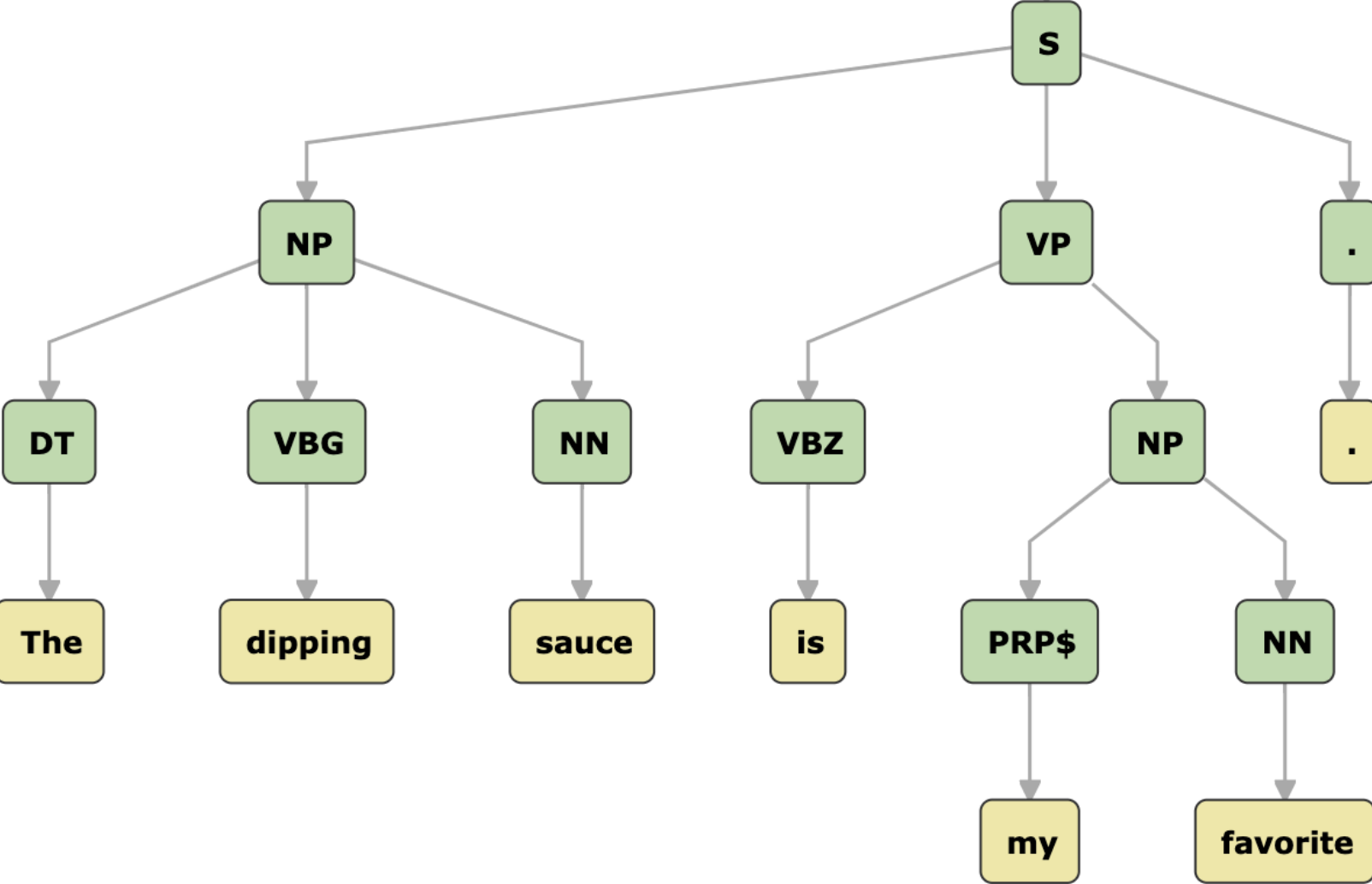}}
	\caption{An example result of constituency analysis.}
	\label{fig:constituency}
\end{figure}

\noindent\textbf{Constituency Analysis.} 
Dependency parsing can only recognize phrases when adjectives or adverbs are present.
However, not all opinion phrases contain adjectives, such as the sentence ``the dipping sauce is my favorite''.
Therefore, we additionally propose the constituency analysis module to make up the shortage.
This module identifies a \textit{noun} subject and its consecutive \textit{verb} phrases as a candidate phrase. 
The constituency parser outputs consecutive multi-word sequences as different components of a sentence, which is illustrated in Fig~\ref{fig:constituency}.
The whole sentence is a parsing tree, where ``the dipping sauce'' is a root-level noun phrase (NP), ``is my favourite'' is labeled as its root-level consecutive verb phrase (VP).
The shorter sequence ``my favourite'' is a leaf-level noun phrase.
To obtain a complete opinion phrase, we take all \textit{root-level} noun phrases and their consecutive verb phrases as candidate opinion phrases.

\subsection{Datasets and Experimental Settings}
\begin{table}[t]
	\centering
	\resizebox{0.47\textwidth}{!}{
	\begin{tabular}{|c|c|c|}
    \hline
    \multirow{5}{*}{\tabincell{c}{Restaurant\\ Aspect (643)}} & Location (5) & street block river avenue \\
    \cline{2-3}
     &Drinks (25) & beverage wines cocktail sake\\
    \cline{2-3}
     &Food (345) & spicy sushi pizza taste\\
    \cline{2-3}
     &Ambience (67)& atmosphere room seating environment\\
    \cline{2-3}
     &Service (201)& tips manager waitress servers\\
     \hline
    \multirow{8}{*}{\tabincell{c}{Laptop\\ Aspect (307)}} & Support (34)& service warranty coverage replace \\
    \cline{2-3}
     &OS (42)& windows ios mac system\\
    \cline{2-3}
     &Display (59)& screen led monitor resolution\\
    \cline{2-3}
     &Battery (38)& life charge last power\\
    \cline{2-3}
     &Company (47)& hp toshiba dell lenovo\\
    \cline{2-3}
     &Mouse (35)& touch track button pad\\
    \cline{2-3}
     &Software (20)& programs apps itunes photoshop\\
    \cline{2-3}
     &Keyboard (32)& key space type keys\\
    \hline
     \multirow{2}{*}{\tabincell{c}{Restaurant\\ Sentiment (643)}} & Good (403) & great nice excellent perfect \\
    \cline{2-3}
     &Bad (240) & terrible horrible disappointed awful\\
    \hline
    \multirow{2}{*}{\tabincell{c}{Laptop\\ Sentiment (307)}} & Good (150)& great nice excellent perfect \\
    \cline{2-3}
     &Bad (157)& terrible horrible disappointed awful\\
    \hline
	\end{tabular}}
	\caption{Keywords and statistics of each aspect/sentiment.}
	\label{tab:dataset_keyword}
\end{table}
\begin{table}[t]
	\centering
	\resizebox{0.48\textwidth}{!}{
	\begin{tabular}{|c|cccc|}
    \hline
	\textbf{Phrase}
            & \tabincell{c}{\textbf{FineSum}\\\textbf{w/o BERT}} & \tabincell{c}{\textbf{FineSum}\\\textbf{w/o joint}} & \textbf{FineSum} & \textbf{Truth}\\
			\hline
			 hot tucson dog place&location &food &food & food \\
			\hline
			 bloody delicious marys&drink &food &drink & drink\\
			\hline
			 find everything ok&food &service &none&none \\
			\hline
			\tabincell{c}{we went yesterday on taco\\ tuesday to meet friends}&food &location &none &none\\
			\hline
			wrong food order&food &food &food &none\\
			\hline
			friendly environment&none &service &none &ambience\\
			\hline
	\end{tabular}}
	\caption{Comparison of predictions on sample phrases between FineSum and its variants.}
	\label{tab:phrase_pred_case}
\end{table}
\subsubsection{Datasets}
Table~\ref{tab:dataset_keyword} displays seed keywords for each aspect and sentiment.
The seed keywords along with aspect names are used as the only supervision for FineSum.
We also display the number of samples under each aspect and sentiment.

\subsubsection{Experimental Settings}
We first explain how we select hyperparameters: 
We set the probability thresholds $\theta_1=0.35$, $\theta_2=0.30$ because it shows the best performances on our self-annotated phrase validation set.
We set the context window length $h=5$, embedding dimension$=100$ by following the settings in the hierarchical topic mining paper~\cite{meng2020hierarchical}.
We adopt the default learning rate $=1e-5$ in the Hugging Face transformer package\footnote{https://huggingface.co/transformers/}, and set batch size$=64$ according to the maximum capacity of our computing facilities.
When selecting the top-$K$ sentences as training data, we initially set $K=2000$ and observed promising results, so we did not try a larger value of $K$.
We conducted all model training using a single NVIDIA GTX 1080 Ti.

\subsection{Evaluation on Phrase Classification}
\subsubsection{Evaluation Details}
\label{sec:phrase}
In the phrase annotation process, we ask three human annotators to assign aspect labels to 500 randomly extracted restaurant phrases.
Annotators consist of two graduates and one undergraduate, all with CS background (as labeling restaurant aspects requires little domain knowledge).
Considering there exist non- and multi-aspects, we define the agreement among annotators as: Given a phrase, all three annotators provide the same number(s) and categor(ies) of aspect(s).

\subsubsection{Qualitative Evaluation}
We qualitatively compare FineSum with its two variants and show results in Table~\ref{tab:phrase_pred_case}.
From the first two rows, we observe that FineSum is able to correct the false prediction from either opinion-oriented embedding (FineSum w/o BERT) and sentence-level BERT classifier (FineSum w/o joint).
The third and fourth rows show that FineSum can distinguish non-aspect phrases better than its variants, as it is fine-tuned to output uniform aspect distribution on non-aspect phrases.
\begin{table*}[!h]
	\centering
	\resizebox{0.95\textwidth}{!}{
	\begin{tabular}{|l|l|}
    \hline
	\tabincell{c}{\textbf{\centering \qquad\qquad\qquad\qquad\qquad\qquad\qquad Cluster}} & \tabincell{c}{\textbf{\centering \qquad\qquad\quad Intrusion Phrase}}\\
	\hline   
	 \tabincell{l}{the \textbf{\textit{bread}} basket was also something that stood out; pretty nice \textbf{\textit{bread}} basket; the\\ best thing i can say about the restaurant is the \textbf{\textit{bread}} basket; the \textbf{\textit{bread}} basket \\was a unique and unexpected touch; \textbf{\textit{bread}} basket is also a lovely touch} &\tabincell{l}{it ranged from jalapeno corn \textbf{\textit{bread}} to\\ a sweet dessert \textbf{\textit{bread}}}\\
    \hline   
    \tabincell{l}{the lime \textbf{\textit{margarita}} tasted like pure lime; the white peach \textbf{\textit{margarita}} is tasty;\\ skinny delicious and thankfully overly sweet \textbf{\textit{margarita}}; delicious pear \\\textbf{\textit{margarita}}; the \textbf{\textit{margarita}} that i had was tasty too} & \tabincell{l}{everyone had the cactus pear \textbf{\textit{margarita}}}\\
    \hline   
    \tabincell{l}{recommend \textbf{\textit{this}} place for lunch; love \textbf{\textit{this}} restaurant though; i love \textbf{\textit{this}} restaurant;\\ \textbf{\textit{this}} was perfect lol; \textbf{\textit{this}} one is a definite miss for dinner} & we visit \textbf{\textit{this}} one twice\\
    \hline   
	\end{tabular}}
	\caption{Examples of the intrusion test on fine-grained clustering.
	For each opinion cluster, we randomly sample an intrusion phrase from outside the cluster.
	To evaluate the distinctiveness of clusters, we require all in-cluster phrases and the intrusion phrase to share at least one \textbf{\textit{common word}}, which is \textit{italicized} and in \textbf{bold}.
	}
	\label{tab:intrusion_exam}
\end{table*}
The last two rows showcase the false prediction of FineSum. 
It wrongly predicts the phrase ``wrong food order'' into the food aspect, suggesting that FineSum may rely excessively on aspect-indicative words to make prediction.
Future work may focus on improving its performance on these ambiguous and hard phrases.

\subsection{Evaluation on Phrase Clustering}
In this section, we introduce evaluation details of phrase clustering.
We employ intrusion test to evaluate the coherence of opinion clusters.
Table~\ref{tab:intrusion_exam} shows three sample intrusion sets.
During human evaluation, we shuffle the intrusion phrase with its corresponding in-cluster phrases and ask human annotators to identify the intrusion one.
We employ the same annotators as in Section~\ref{sec:phrase}.
For each embedding method, we generate 40 intrusion sets and randomly shuffle sets from different methods during human evaluation.
In total, annotators are asked to work on 3*40 clusters, each containing 6 phrases. 

\bibliography{aaai22}